\documentclass[12pt]{article}
\usepackage{authblk}
\usepackage{amsmath,amsthm,amssymb,amsfonts}
\usepackage{psfrag,epsf}
\usepackage{enumitem}
\usepackage{natbib}
\usepackage{times}
\usepackage{threeparttable}
\usepackage{booktabs}
\usepackage{bm}
\usepackage{xcolor}
\usepackage{threeparttable}
\usepackage{appendix}
\usepackage[scr=rsfs]{mathalpha}
\usepackage{graphicx,subfigure,caption}
\usepackage{algorithm,algorithmic}
\usepackage{multirow}
\usepackage[colorlinks,
linkcolor=red,
anchorcolor=blue,
citecolor=blue]{hyperref}

\newtheorem{theorem}{\textbf{Theorem}}

\newtheorem{corollary}{Corollary}[section]
\newtheorem{lemma}{Lemma}

\newcommand{\y}{\bm{y}}
\newcommand{\X}{\bm{X}}

\newcommand{\MM}{\widehat{\mathcal{M}}}

\begin{document}

\def\spacingset#1{\renewcommand{\baselinestretch}%
{#1}\small\normalsize} \spacingset{1}

\addtolength{\textheight}{.5in}%
%%%%%%%%%%%%%%%%%%%%%%%%%%%%%%%%%%%%%%%%%%%%%%%%%%%%%%%%%%%%%%%%%%%%%%%%%%%%%%

\title{\bf 
An adaptive subsampling method for large-sample feature screening
% Adaptive Feature Screening via Online Best-Arm Identification
}

\author{Xiaxue Ouyang}
\author[1,*]{Kejun He}
\author[1,*]{Cheng Meng}

\affil[1]{Institute of Statistics and Big Data, Renmin University of China, Beijing, China}
\affil[*]{Corresponding author: Kejun He, kejunhe@ruc.edu.cn}
\affil[*]{Corresponding author: Cheng Meng, chengmeng@ruc.edu.cn}
% \author{}
\date{}
\maketitle

\spacingset{1.5}
\begin{abstract}
We consider the sure independence screening (SIS) method, a standard feature screening approach that aims to eliminate non-informative features in ultrahigh-dimensional datasets.
Although effective, SIS incurs a computational cost of order $O(np)$ for a predictor matrix of size $n \times p$, which can be prohibitively expensive when both $n$ and $p$ are considerable.
Motivated by the multi-armed bandit (MAB) problem, we propose a more computationally efficient feature screening algorithm that reduces the cost to $O(\sqrt{n}p)$.
The core idea is to progressively increase the subsample size and eliminate variables with small empirical marginal Pearson correlations, thereby avoiding unnecessary computation on unpromising features.
We develop a new interpretable statistical theoretical analysis that characterizes how the subsample size affects screening accuracy, thereby revealing the balance between computational efficiency and statistical reliability. Moreover, we show that the proposed method retains the sure screening property under mild regularity conditions.
Extensive numerical experiments on synthetic and real-world datasets show that BanditSIS achieves screening and prediction performance comparable to SIS while substantially reducing computational time.
Our method offers a scalable and adaptive alternative to SIS, particularly well-suited for large-sample, high-dimensional applications where computational efficiency is critical.
\end{abstract}
	
\noindent%

% Multi-armed bandits; Online learning; Best-arm identification; Linear Bandit; Adaptive subsampling
	
\spacingset{1.45}
%%%%%%%%%%%%%%%%%%%%%%%%%%%%%%%%%%%%%%%%%%%%%%%%%%%%%%%%%%%%%%%%%%%%%%%%%%%%%%%%%%%%%%%%%%
%                                                                                        %
%                                                                                        %
%                                                                                        %
%                                                                                        %
%                                                                                        %
%                                                                                        %
%                                     INTRODUCTION                                       %
%                                                                                        %
%                                                                                        %
%                                                                                        %
%                                                                                        %
%                                                                                        %
%                                                                                        %
%%%%%%%%%%%%%%%%%%%%%%%%%%%%%%%%%%%%%%%%%%%%%%%%%%%%%%%%%%%%%%%%%%%%%%%%%%%%%%%%%%%%%%%%%%
\section{Introduction}\label{sec:intro}
With the rapid advancement of science and technology, high-dimensional data have become increasingly prevalent across a broad range of fields. The availability of such data creates unprecedented opportunities for scientific discovery and data-driven decision-making. At the same time, however, the analysis of high-dimensional datasets poses substantial challenges to conventional statistical and machine learning methods, primarily due to the considerable computational and memory demands involved.

To address the difficulties associated with high dimensionality, a large body of literature has focused on dimensionality reduction; see, for example, \citet{donoho2000high} and \citet{fan2006statistical}. Broadly speaking, existing approaches can be divided into two major categories: feature projection and feature selection. Projection-based methods map the original data from a high-dimensional space to a lower-dimensional subspace or manifold while preserving important structural  features of the original data. Representative examples include principal component analysis (PCA, \citealt{hotelling1992relations}), linear discriminant analysis (LDA, \citealt{fisher1936use}), multidimensional scaling (MDS, \citealt{mead1992review}; \citealt{borg2003modern}), locally linear embedding (LLE, \citealt{roweis2000nonlinear}), local tangent space alignment (LTSA, \citealt{zhang2004principal}), and t-distributed stochastic neighbor embedding (t-SNE, \citealt{hinton2002stochastic}). 
In contrast, feature selection seeks to identify a subset of features that are most relevant for model construction and interpretation. Such methods include LASSO \citep{tibshirani1996regression}, smoothly clipped absolute deviation (SCAD, \citealt{fan2001variable}), least angle regression (LARS, \citealt{efron2004least}), the Dantzig selector \citep{candes2007dantzig}, adaptive LASSO \citep{zou2006adaptive}, and the elastic net \citep{zou2005regularization,zou2009adaptive}. 
In some particular application domains, feature selection offers clear advantages over feature projection because it directly isolates informative raw variables. For instance, \citet{liu2018feature} noted that in biomedical studies, feature selection methods can be used to identify disease-related genes from gene expression data, thereby facilitating the development of targeted therapeutic strategies.

In recent decades, developments in areas such as genomics, neural network modeling, and large-scale user information collection have led to an explosive increase in data dimensionality, often at a rate that may grow exponentially with the sample size. As emphasized by \citet{fan2009ultrahigh}, such ultrahigh-dimensional settings pose serious difficulties for classical feature selection methods in terms of computational feasibility, statistical accuracy, and algorithmic stability. 
A widely used strategy for mitigating these issues is sure independence screening (SIS), proposed by \citet{fan2008sure}. The main goal of SIS is to reduce the number of candidate variables from an extremely large scale to a moderate size by screening out irrelevant features in an initial preprocessing step. In linear models, SIS ranks predictors according to their marginal Pearson correlations with the response, attempting to retain all the important features in the reduced feature space. Its central theoretical guarantee is the sure screening property, namely, that the reduced feature set contains all truly important variables with probability tending to one. Despite its effectiveness, SIS may still be computationally demanding in large-scale applications. For a predictor matrix of size $n \times p$, its computational cost is of order $O(np)$. This burden becomes particularly restrictive when both $n$ and $p$ are large, especially in modern applications where the sample size may reach the order of millions.

To alleviate this computational bottleneck, we propose a new algorithm for large-sample SIS, termed BanditSIS. Our approach is motivated by the multi-armed bandit (MAB) framework \citep{berry1985bandit, chakrabarti2008mortal}, and in particular by the perspective of adaptive resource allocation under limited computational budgets. In the classical best-arm identification problem, one is given $p$ arms, each associated with an unknown reward distribution, and the goal is to identify the arm with the largest expected reward using as few samples as possible. Inspired by this paradigm, we reinterpret feature screening as a best-arm identification problem: each feature is viewed as an arm, its marginal correlation plays the role of a reward proxy, and the target set of $d$ informative predictors corresponds to the top $d$ best arms. Based on this formulation, BanditSIS employs an adaptive subsampling and elimination mechanism that progressively discards features with relatively small estimated marginal correlations, thereby concentrating computational effort on promising predictors and substantially reducing the overall screening cost.

Our contributions are summarized as follows. 
\begin{itemize}
    \item We propose BanditSIS, a bandit-inspired adaptive subsampling algorithm for large-sample sure independence screening. Instead of computing marginal correlations on the full \(n\times p\) data matrix as in SIS, BanditSIS substantially reduces unnecessary computations on unpromising features and lowers the computational cost from \(O(np)\) to \(O(\sqrt n p)\).
    \item We develop a new interpretable theoretical analysis tailored to the adaptive subsampling mechanism. We explicitly characterize how the subsample size and the feature dimension jointly determine the screening accuracy, explaining how computational savings and screening accuracy are balanced across rounds. Also, we establish that the proposed method preserves the desired sure screening property.
    \item Extensive simulations and a real-data analysis demonstrate that BanditSIS achieves screening and prediction performance comparable to SIS while requiring substantially less computational time.
\end{itemize}

The rest of this article is organized as follows. In Section \ref{sec:background}, we provide a brief review of SIS and its theoretical properties, along with the motivation for our work based on the MAB problem. In Section \ref{sec:method}, we develop a more computationally efficient screening approach compared to SIS and give the corresponding algorithm by integrating insights from the MAB framework into SIS. 
The theoretical properties of our approach are provided in Section \ref{sec: theory}. In Section \ref{sec:verify}, we illustrate the performance of our method and compare it with SIS through simulation and a real dataset. Section \ref{sec:conclusion} concludes the paper and the technical proofs are given in the Supplementary Material.

%%%%%%%%%%%%%%%%%%%%%%%%%%%%%%%%%%%%%%%%%%%%%%%%%%%%%%%%%%%%%%%%%%%%%%%%%%%%%%%%%%%%%%%%%%
%                                                                                        %
%                                                                                        %
%                                                                                        %
%                                                                                        %
%                                                                                        %
%                                                                                        %
%                                     Section 2                                          %
%                                                                                        %
%                                                                                        %
%                                                                                        %
%                                                                                        %
%                                                                                        %
%                                                                                        %
%%%%%%%%%%%%%%%%%%%%%%%%%%%%%%%%%%%%%%%%%%%%%%%%%%%%%%%%%%%%%%%%%%%%%%%%%%%%%%%%%%%%%%%%%%
	
\section{Background} \label{sec:background}
\subsection{Sure Independence Screening}
\label{sec:background-SIS}
Following the notations in the literature \citet{fan2008sure}, we consider a multivariate linear model, with $n$ independent samples and $p$ features,
\begin{equation}\label{eqn:linear_model}
    \y = \X\bm{\beta}+ \bm{\epsilon},
\end{equation}
where $\y = (y_1,y_2,\ldots,y_n)^\top$ is the response vector, $\X = (\bm{x}_1,\bm{x}_2,\ldots,\bm{x}_n)^\top=(X_1,X_2,\ldots,X_p)$ is an $n \times p$ random design matrix, $\bm{\epsilon} = (\epsilon_1,\epsilon_2,\ldots,\epsilon_n)^\top$ is an $n$-vector of i.i.d. random errors, and $\bm{\beta} = (\beta_1,\beta_2,\ldots,\beta_p)^\top$ is the coefficient vector. For large $p$, the basic assumption is that only a small number of predictors among $X_1,\ldots,X_p$ contribute to the response $\y$. This corresponds to the sparse assumption of the coefficient vector $\bm{\beta}$. Denote $\mu_j = \mathbb{E}(X_j), \mu_Y = \mathbb{E}(Y),  \sigma_j^2 = \operatorname{var}(X_j), \sigma_Y^2 = \operatorname{var}(Y)$,
and $\sigma_{jY} = \operatorname{cov}(X_j,Y)$.

Intuitively, the idea of SIS is to independently compute the marginal Pearson correlation between each predictor and the response to evaluate how much contribution it makes to the response and then retain the features with the larger absolute value of the correlation coefficient and drop the weaker ones. 
In particular, the standard pipeline of SIS involves the following steps.
First, we standardize each input predictor and the response so that the observed mean is zero and the sample standard deviation is one. 
Denote $\widehat{\bm\omega}=(\widehat{\omega}_1,\widehat{\omega}_2,\ldots,\widehat{\omega}_p)^\top$ as a vector of marginal Pearson coefficients of the standardized predictors $X_1,\ldots,X_p$ with the response $\y$, i.e., $\widehat{\bm\omega} = \X^\top \y / n.$
%The marginal Pearson coefficient $\omega_j$ measures the linear dependency between $X_j$ and $\y$, for any $j=1,2,\ldots,p$.
%The stronger the linear correlation between $X_j$ and $\y$ is, the larger the absolute value of the coefficient $\omega_j$ will be. 
The population Pearson correlation is $ \omega_j = \sigma_{jY} / \sigma_j\sigma_Y$.
Next, we sort the $p$ componentwise magnitudes of $\widehat{\bm\omega}$ in decreasing order and put the features corresponding to the first $\lfloor\gamma n\rfloor$ largest components into a set $\mathcal{M}_{\gamma}$, i.e.,
\begin{equation*}
	 \MM_{\gamma} = \left\{ 1 \leq j \leq p:|\widehat{\omega}_j| \text{ is among the first } \lfloor\gamma n\rfloor \text{ largest of all}\right \},
 \end{equation*}
where $\gamma$ is a given constant from 0 to 1 and $\lfloor\gamma n\rfloor$ is the integer part of $\gamma n$. 
Finally, $\mathcal{M}_{\gamma}$ is the submodel that we have filtered out with size $d=\lfloor\gamma n\rfloor < n$ from the full model $\left\{ 1,2,\ldots,p \right\}$.
%We can reduce the $p$ features to a given size $d < n$, for instance, usually $n-1$ or $n/\log(n)$.

It is well-known that SIS enjoys the so-called sure screening property.
Specifically, let $\mathcal{M}_*=\{1 \leq j \leq p : \beta_j \ne 0 \}$ be the set that contains all important variables. 
It can be shown that when $\gamma$ satisfies a certain order, all important variables will be included in the selected submodel $\mathcal{M}_\gamma$ with asymptotic probability 1 as $n$ diverges to infinity, i.e.,
\begin{equation}\label{eqn:sure_screening}
	 \mathbb{P} \left\{ \mathcal{M}_* \subset \MM_{\gamma} \right\} \rightarrow 1 \text{, as } n \rightarrow \infty.
\end{equation}

\subsection{Motivation}\label{sec:background-MAB}

The MAB problem is one of the classical problems in decision theory and control. It involves a trade-off between exploration (making decisions to acquire unknown information) and exploitation (using the known information to maximize the current return). In MAB, there are several arms, each of which corresponds to an unknown distribution of rewards. For each pull of an arm, it will return a reward generated independently and randomly from the corresponding unknown reward distribution. The goal of MAB, illustrated by \citet{chen2014combinatorial} and \citet{kaufmann2016complexity}, is to (1) obtain as many rewards as possible within a limited number of pulls, or (2) identify the best arm with the highest expected reward by using the fewest number of pulls. 
In this paper, we are concerned with the latter. In general, the problem of best arm identification has two different stopping conditions \citep{gabillon2012best,tran2010epsilon,ding2013multi,jamieson2014best}:
\begin{itemize}
    \item \textit{Fixed budget}. With a fixed computational budget, the MAB algorithm stops once it reaches the predetermined computational cost. Then it returns the arm which approximates the true best arm as closely as possible.
    \item \textit{Fixed confidence}. In fixed confidence settings, the MAB algorithm seeks to minimize the number of pulls used while guaranteeing that it returns an arm with a certain level of accuracy with a certain probability.
\end{itemize}

Motivated by the second goal, we propose an adaptive subsampling strategy that regards candidate features as arms and the informative features as the best arms. 
In this framework, the arm parameters correspond to the correlation coefficients, and pulling an arm corresponds to computing the marginal Pearson correlation on randomly selected subsamples. 
Our goal is to screen out informative features using substantially fewer observations than SIS, following the spirit of best-arm identification algorithms.
We propose the BanditSIS algorithm, which adjusts the number of observations used in each iteration to improve computational efficiency while preserving a certain screening accuracy.
The detailed procedure of the proposed algorithm is presented below.
%%%%%%%%%%%%%%%%%%%%%%%%%%%%%%%%%%%%%%%%%%%%%%%%%%%%%%%%%%%%%%%%%%%%%%%%%%%%%%%%%%%%%%%%%%
%                                                                                        %
%                                                                                        %
%                                                                                        %
%                                                                                        %
%                                                                                        %
%                                                                                        %
%                                     Section 3                                          %
%                                                                                        %
%                                                                                        %
%                                                                                        %
%                                                                                        %
%                                                                                        %
%                                                                                        %
%%%%%%%%%%%%%%%%%%%%%%%%%%%%%%%%%%%%%%%%%%%%%%%%%%%%%%%%%%%%%%%%%%%%%%%%%%%%%%%%%%%%%%%%%%

\section{Method}\label{sec:method}
Our method adopts an iterative strategy. At each step, it randomly subsamples the data, computes marginal correlation coefficients, and discards the least informative features.
Let $p_l$ denote the number of remaining features at the end of the $l$th round.
% To control the number of retained features, we define a shrinking parameter $0<\delta_l<1$ such that $p_l = \lfloor\delta_l p_{l-1}\rfloor$.
We employ a modified version of the median elimination strategy to determine the number of retained features. 
The basic idea of the median elimination strategy is to discard the bottom half of the candidates in each iteration. 
This strategy is widely used in the best arm identification problem due to its simplicity and effectiveness \citep{even2006action,liu2019bandit}. 
By avoiding unnecessary explorations of poorly performing arms, it allocates more computational resources to promising ones and reduces the cost of exploration, accelerating the entire learning process of the algorithm.
In our proposed algorithm, we extend the classical median elimination strategy to obtain the top $d$ features rather than a single best. Specifically, at the end of the $l$th iteration, we discard $\lceil (p_{l-1}-d)/2 \rceil$ features with the smallest absolute marginal correlations and retain the top $\lfloor (p_{l-1} + d)/2 \rfloor$ features in the submodel $\MM_l$. 
% This implies that $\delta_l = (p_{l-1}+d)/(2p_{l-1})$.

We now discuss how to determine the number of subsamples $n_l$ used in the $l$th round. 
Our objective is to construct an efficient variant of the SIS method such that it still preserves the sure screening property described in Equation~(\ref{eqn:sure_screening}) under the median elimination strategy.
To achieve this goal, we propose a function $t(n,\alpha_{l-1})$ to determine each $n_l$.
In particular, we set a hyperparameter $\alpha>0$, let $\alpha_l$ = $\alpha/1.1^l$ for $l=1,2,\ldots,k$, $\alpha_0 = \alpha$, and
\begin{equation}\label{eqn:nl}
    n_l = t(n,\alpha_{l-1}) := \frac{n(\alpha_{l-1}^2+1)}{\alpha_{l-1}^2\sqrt{n}+1}.
\end{equation}
The hyperparameter $\alpha$ here controls the trade-off between the screening efficiency and accuracy. 
In the next section, we show that this proposed functional form $t(n,\alpha)$ enjoys the following properties: (1) it is compatible with the median elimination strategy, (2) it preserves the sure screening property, and (3) it results in a much lower computational complexity compared to SIS.

\begin{figure}
    \centering
    \includegraphics[width=0.9\linewidth]{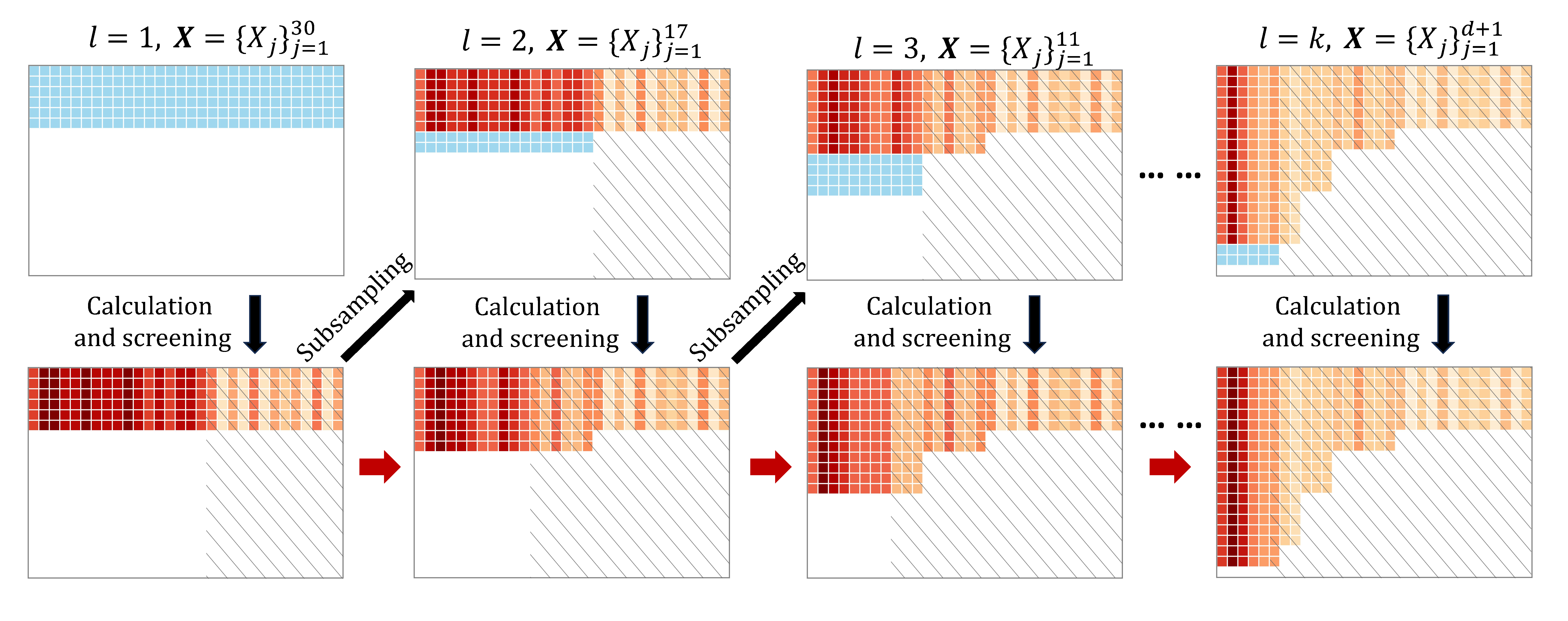}
    \caption{An illustration of the proposed BanditSIS algorithm. 
    % a) Randomly shuffle the full sample (the predictor features and the corresponding response). (
    For clearer visualization, the data are randomly shuffled in advance, so that the subsamples selected at each round can be displayed sequentially. 
    The light blue entries represent the subsamples (rows) selected in the current round. 
    The selected entries in the predictor matrix are labeled with red colors, where darker red corresponds to a larger absolute magnitude of the marginal correlation.
    The shaded entries represent the predictor features (columns) that have been discarded in the current and the previous rounds. 
    The proposed algorithm alternatively drops the features with low absolute marginal correlation and selects subsamples in each round.}
    \label{fig:alg}
\end{figure}

Intuitively, our BanditSIS algorithm can be summarized in the following steps.
% First, we randomly shuffle the entire dataset. This simple pre-processing step reduces memory access overhead and minimizes data replication in subsequent computation stages, significantly improving practical efficiency. 
First, initialize $\MM_0$ as the full model $\{ 1,2,\ldots,p\}$, and $p_0 =p$.
In the first round, we calculate the marginal correlation using $n_1$ randomly selected subsamples, where $j \in \MM_0$, and obtain a submodel $\MM_1$ with size $p_1=\lfloor (p_0 + d)/2 \rfloor$.
We then apply a similar procedure to the model $\MM_1$ using $n_2$ randomly selected subsamples in total, obtaining a submodel $\MM_2 \subset \MM_1$ with size $p_2 = \lfloor (p_1 + d)/2 \rfloor$.
Repeat these steps until the $k$th round such that the model size reaches the target size $d$, i.e., $|\MM_{k-1}| = p_{k-1} > d$, and $|\MM_k| = p_k = d$.
% For computational concern, in practice, instead of randomly selecting subsamples in each round, we first randomly shuffle the full sample $\{ (\bm{x}_i,y_i) \}^{n}_{i=1}$, i.e., the predictor features and the corresponding response, and then sequentially select the subsample in each round.  
The BanditSIS algorithm is summarized in Algorithm~\ref{algm:BanditSIS} and is illustrated in Fig.~\ref{fig:alg}.

\begin{algorithm}[t]
\caption{BanditSIS}
\label{algm:BanditSIS}
\begin{algorithmic}[1]
   \STATE {\bf Input:} Data $\{ (\bm{x}_i,y_i) \}^{n}_{i=1}$, model size $d$, hyperparameter $\alpha$ 
   \STATE {\bf Initialize:} $\mathcal{M} = \{ 1,2,\ldots,p\}$, $n_0=0$, $l=1$
   % \STATE Randomly shuffle the data $\{ (\bm{x}_i,y_i) \}^{n}_{i=1}$
   \WHILE { $|\mathcal{M}| > d$}
		\STATE Calculate $n_l = t(n,\alpha)$ and select $n_l - n_{l-1}$ subsamples sequentially from feature $X_j$, where $j \in \mathcal{M}$
		\STATE Update the correlation coefficient $\omega_j$ with the selected subsamples
		\STATE $\mathcal{M} \leftarrow \{ j \in \mathcal{M}: |\omega_j| \text{ is among the first } \Big\lfloor \frac{|\mathcal{M}| + d}{2} \Big\rfloor \text{\quad largest of all} \}$
		\STATE $l \leftarrow l+1$; $\alpha \leftarrow \alpha/1.1$
   \ENDWHILE
   \STATE {\bf Return:} $\MM_B \leftarrow \mathcal{M}$
\end{algorithmic}
\end{algorithm}

The overall computational complexity of BanditSIS is $O(\sqrt{n}p)$. 
Specifically, the computational complexity of BanditSIS, which iteratively updates the correlation coefficients through the selected subsamples, is $O(\sqrt{n}p/\alpha^2)$.
% This cost arises from two primary components of the algorithm. Specifically, the first step of BanditSIS, which involves randomly shuffling the entire dataset, incurs a computational complexity of $O(n\log{n})$. This simple pre-processing step reduces memory access overhead and minimizes data replication in subsequent computation stages, significantly improving practical efficiency. 
% The second component of computational complexity, which iteratively updates the correlation coefficients through the selected subsamples, is $O(\sqrt{n}p/\alpha^2)$. 
Such complexity is lower than that of SIS since we alternate between sequentially selecting subsamples and dropping the least informative features. By progressively focusing on the most promising predictors, the algorithm avoids unnecessary computation on the irrelevant ones.
% In particular, this part of computational complexity is related to the total number of iterations $k$, which is on the order of $\log(p)$.
% This precisely because the algorithm proceeds iteratively, and in each round, only the top $\lfloor (p_{l-1} + d)/2 \rfloor$ most promising features out of $p_{l-1}$ features will remain until only $d$ features survive. This process inevitably results in the $\log(p)$ term in the computational complexity.
It is important to note that the precise complexity also depends on the algorithm hyperparameter $\alpha$, which controls the size of subsamples. For the sake of brevity, we omit the multiplicative factor in the main expression.
% , but the complexity bound includes an additional multiplicative factor of $O(1/\alpha_0^2)$.
The technical proof of computational complexity is provided in the Supplementary Material.

%%%%%%%%%%%%%%%%%%%%%%%%%%%%%%%%%%%%%%%%%%%%%%%%%%%%%%%%%%%%%%%%%%%%%%%%%%%%%%%%%%%%%%%%%%
%                                                                                        %
%                                                                                        %
%                                                                                        %
%                                                                                        %
%                                                                                        %
%                                                                                        %
%                                     Section 4                                          %
%                                                                                        %
%                                                                                        %
%                                                                                        %
%                                                                                        %
%                                                                                        %
%                                                                                        %
%%%%%%%%%%%%%%%%%%%%%%%%%%%%%%%%%%%%%%%%%%%%%%%%%%%%%%%%%%%%%%%%%%%%%%%%%%%%%%%%%%%%%%%%%%
\section{Theoretical Results} \label{sec: theory}
In this section, we show that the proposed BanditSIS algorithm preserves the sure screening property under mild regularity conditions.
Compared to the assumptions imposed in the original SIS theory \citep{fan2008sure}, our analysis replaces the global design-matrix concentration and instead relies on the concentration of marginal utilities together with a round-wise separation condition.
These conditions are commonly encountered in marginal screening literature. 
In addition, such regularity conditions introduced are mainly used to establish the theoretical guarantee and to clarify the statistical mechanism behind the proposed adaptive subsampling procedure. 
They are not intended to be the weakest possible assumptions. The technical proofs are relegated to Supplementary Material.

% 需要对 Δ_l 进行补充（条件比之前更弱了），Δ_l 是什么意思？条件弱在了哪里？
% 取不同的阶数/常数，可以导出对 p 的阶数要求也不同 （算了没必要讨论）
% 其实我们是把集中不等式 / sure screening property 重新证明了一遍的（贡献提一下就行）
% 如果我们的条件和 SIS 一样，也是可以推出来我们的结论的。可以在附录里面讨论条件弱化的对比（算了感觉也没必要讨论）
% Condition 3其实以及蕴含着对共线性的假设，我们通过Condition 3，完全绕开了对beta_j的讨论和共线性的讨论，而把目光完全 focus on 边际的 Pearson correlation 上，下面是对共线性的一个讨论，有需求再补充上去
% In the linear model, the population marginal covariance satisfies $\operatorname{cov}(X_j,Y)=(\Sigma\beta)_j,$ rather than being determined by \(\beta_j\) alone. Therefore, collinearity can either attenuate the marginal utilities of active variables or inflate those of inactive variables. Our analysis does not separately impose an eigenvalue condition on \(\Sigma\); instead, its joint effect with the coefficient vector is summarized through the round-wise marginal gap \(\Delta_l\). If collinearity makes an active variable marginally indistinguishable from the oracle-discarded variables, then \(\Delta_l\) becomes small and the theoretical guarantee correspondingly weakens.
\begin{itemize}
    \item Condition 1. $\log(p) \leq C_pn^\xi$ for some constants $C_p>0$ and $\xi \in (0,1/2-2\kappa)$, where $\kappa$ is given by Condition 4.
    \item  Condition 2. The predictors $\{X_j\}_{j=1}^p$ and the response $Y$ are sub-Gaussian variables, and there exist constants $0<c_\sigma<C_\sigma<\infty$ such that, uniformly over $j$,
    $c_\sigma \le \sigma_j^2, \sigma_Y^2 \le C_\sigma$.
    \item Condition 3. Define the oracle retained set
    \begin{equation*}
        \mathcal{M}_l^\circ (\mathcal{A}) =
        \left\{ j \in \mathcal{A}:
        |\omega_j| \text{ is among the top } \left\lfloor  \frac{|\mathcal{A}| + d}{2} \right\rfloor
        \text{ largest values}\right\}.
    \end{equation*}
    Assume that there exists a deterministic sequence $\Delta_l>0$, $l=1,\dots,k$, such that, for any possible retained set $\mathcal{A} \subset \{1, \ldots, p\}$ satisfying $\mathcal{M}^\ast \subset \mathcal{A}$ and $|\mathcal{A}| = p_{l-1}$, 
    \begin{equation*}
        \min_{j \in \mathcal{M}^\ast} |\omega_j|
        -
        \max_{h \in \mathcal{A} \setminus \mathcal{M}_l^\circ(\mathcal{A})} |\omega_h|
        \ge
        \Delta_l .
    \end{equation*}
    \item Condition 4. For some constants $0 \leq \kappa < 1/4$ and $C_\Delta > 0$
    \begin{equation*}
        \Delta_{\min} = \min_{1 \leq l \leq k}\Delta_l \geq C_\Delta n^{-\kappa}.
    \end{equation*}
\end{itemize}

These conditions are parallel to the regularity conditions in SIS theory \citep{fan2008sure}.
Condition 1 specifies the admissible dimensional growth regime, especially allowing for the ultrahigh dimensional setting that even the number of features $p$ grows nearly exponentially with a power of $n$.
This high-dimensional growth condition is mainly theoretical, ensuring that the sure screening property still holds even in this challenging regime. 
Condition 2 is imposed to guarantee concentration of the empirical marginal utilities. The sub-Gaussian condition ensures concentration of empirical covariance and variance terms, while the lower variance bound prevents the denominator of the Pearson correlation from degenerating. 
Condition 3 is a round-wise population separation condition, which requires the active variables to be separated from those that would be discarded by an oracle procedure using the population marginal utilities. 
It only requires active variables to be separated from those that would be removed by the oracle elimination rule at each round, instead of requiring active variables to dominate all inactive variables in the population marginal ranking. 
This is weaker than a global separation condition \citep{cui2015model,li2023scalable}.
Condition 4 controls the weakest round-wise separation margin, which prevents this gap between between the active variables and the oracle discarded ones from vanishing too quickly \citep{li2012feature,fan2008sure,li2012robust}. 
It plays a role analogous to the minimum signal strength condition in SIS, but it is formulated in terms of the marginal-correlation gap relevant to the adaptive elimination procedure rather than the regression coefficients.

The following lemma establishes the concentration inequality for the empirical Pearson marginal utility under Condition~2. This result is the key probabilistic ingredient in our analysis, as BanditSIS ranks and eliminates variables according to empirical marginal correlations.

\begin{lemma}[Concentration of Pearson correlation]
\label{lem: pearson_concentration}
    Under Condition 2, there exist constants $C>0$, $c>0$, and $t_0>0$, depending on $c_\sigma,C_\sigma$ and the sub-Gaussian norm bound, such that for all $0<t\le t_0$,
    \begin{equation*}
        \mathbb{P}\left\{
        \big||\widehat{\omega}_j|-|\omega_j|\big|>t
        \right\}
        \le
        C\exp(-c n t^2).
    \end{equation*}
\end{lemma}

We now provide our main theorem that indicates the submodel $\MM_B$ provided by BanditSIS contains all the informative features in the true model $\mathcal{M}_*$ with an overwhelming probability. 

\begin{theorem}[Finite-sample sure screening bound]
\label{thm: sure_screening}
    Under Conditions 2--3, there exist constants $C>0$, $c>0$ depending on $c_\sigma,C_\sigma$ and the sub-Gaussian norm bound, such that
    \begin{equation*}
        \mathbb{P}\left\{ \mathcal{M}^\ast \subset \MM_B \right\}
        \geq
        1 - \sum_{l=1}^{k}
        C p\exp(-c n_l\Delta_l^2).
    \end{equation*}
\end{theorem}

Theorem~\ref{thm: sure_screening} establishes a finite-sample screening bound for BanditSIS. 
It shows that the probability of missing at least one important variable is controlled by the accumulated round-wise error $\sum_{l=1}^k C p \exp(-c n_l\Delta_l^2)$.
Here, $n_l\Delta_l^2$ can be interpreted as the effective signal strength in the $l$th round, and successful elimination requires this quantity to dominate the logarithmic feature complexity $\log(p)$. 
This result reveals the statistical mechanism behind BanditSIS and explains how computational savings and screening accuracy are balanced.
Intuitively, early rounds are typically easier because the algorithm discards variables with small marginal utilities, leading to a relatively large \(\Delta_l\); therefore, a small subsample is sufficient. 
Later rounds are more difficult because the retained variables become more competitive and the gap \(\Delta_l\) may become smaller. Accordingly, BanditSIS increases \(n_l\) to obtain more accurate marginal-correlation estimates. Thus, the adaptive subsample schedule is naturally matched to the changing difficulty of the screening task across rounds.
Under the median elimination strategy, the number of rounds satisfies \(k=O(\log p)\). Although \(k\) may diverge with \(n\), its growth is logarithmic in \(p\), and hence is dominated by the exponential decay of the round-wise error terms under Conditions~1--4. This observation leads to Corollary~\ref{cor: sure_screening}, which establishes the sure screening property of BanditSIS.

\begin{corollary}[Sure screening property]
\label{cor: sure_screening}
    Suppose the Conditions 1--4 hold. Under the defined subsample function $n_l = t(n,\alpha_{l-1})$ and the median elimination strategy, we have the sure screening property of BanditSIS that
    \begin{equation*}
        \mathbb{P} \big\{ \mathcal{M}^* \subseteq \MM_B \big\}
        \rightarrow 1 \quad \text{as } n \rightarrow \infty.
    \end{equation*}
\end{corollary}

%%%%%%%%%%%%%%%%%%%%%%%%%%%%%%%%%%%%%%%%%%%%%%%%%%%%%%%%%%%%%%%%%%%%%%%%%%%%%%%%%%%%%%%%%%
%                                                                                        %
%                                                                                        %
%                                                                                        %
%                                                                                        %
%                                                                                        %
%                                                                                        %
%                                     Section 5                                          %
%                                                                                        %
%                                                                                        %
%                                                                                        %
%                                                                                        %
%                                                                                        %
%                                                                                        %
%%%%%%%%%%%%%%%%%%%%%%%%%%%%%%%%%%%%%%%%%%%%%%%%%%%%%%%%%%%%%%%%%%%%%%%%%%%%%%%%%%%%%%%%%%
\section{Numerical Experiments}
\label{sec:verify}
In this section, we evaluate the numerical performance of BanditSIS through both simulation and real data analysis. 
We also compare the computation time of BanditSIS with that of SIS.

\subsection{Simulation}
Let the predictors $\X = (X_1,X_2,\ldots,X_p)$ generate from a multivariate distribution with mean zero and covariance matrix $\bm{\Sigma} = (\sigma_{ij})_{p \times p}$ with $\sigma_{ij} = 0.5^{|i-j|}$, and the i.i.d. error term $\bm{\epsilon}$ from a specified distribution. 
We fix the sample size $n = 1000$, the number of predictors $p = 2000$, and the number of true active predictors $dt=10$.

Let $\y$ be the response variable and $X_j$ be the $j$th feature of $\X$. 
Following \citet{fan2008sure} and \citet{liu2022model}, we consider the following four data generation settings. 

\begin{itemize}
    \item[\textbf{M1:}] $y = \sum\limits_{j=1} ^{dt} X_j + \epsilon$, where $\X \sim N(\bm{0},\bm{\Sigma})$, $\epsilon \sim N(0,1)$.
    \item[\textbf{M2:}] $y = \sum\limits_{j=1} ^{dt} \beta_j X_j + \epsilon$, where $\X \sim N(\bm{0},\bm{\Sigma})$, $\epsilon \sim N(0,1)$, $\beta_j = (-1)^U(a+|Z|)$, $U \sim B(0.95)$, $a = 4\log{n} / \sqrt{n}$, $Z \sim N(0,1)$.
    \item[\textbf{M3:}] $y = \sum\limits_{j=1} ^{dt} X_j + \epsilon$, where $\X \sim t_5(\bm{0},\bm{\Sigma})$, $\epsilon \sim N(0,1)$.
    \item[\textbf{M4:}] $y = \sum\limits_{j=1} ^{dt} X_j + \epsilon$, where $\X \sim 0.5N(\bm{0},\bm{\Sigma}) + 0.5N(\bm{2},\bm{\Sigma})$, $\epsilon \sim N(0,1)$.
\end{itemize}
Here $t_5(\bm{0},\bm{\Sigma})$ represents the multivariate t-distribution with zero mean and degree-of-freedom 5, and its variance-covariance matrix is $\bm{\Sigma}$.
\textbf{M3} is included as a robustness check beyond the sub-Gaussian setting considered in the theory.
%We also consider the scenario where the distribution of $\X$ is relatively heavy-tailed and mixed multivariate normal.
We replicate each experiment 500 times and measure the performance of the proposed method by the following two criteria:
\begin{enumerate}
    \item $\mathcal{S}$: the 5\%, 25\%, 50\%, 75\%, and 95\% quantiles of the minimum model size that includes all active predictors.
    \item $\mathcal{P}$: the proportion that all active predictors are selected for a given model size $d$ over 500 replications. 
\end{enumerate}

\begin{table}[htp]
\centering
\caption{\label{tab:res}The quantiles of minimum model size $\mathcal{S}$ and the proportion $\mathcal{P}$ with $d_1= 144$, $d_2=288$, and $d_3=432$ for settings $\textbf{M1}$--$\textbf{M4}$ over 500 replications. The values of hyperparameter $\alpha$ in BanditSIS are various among 1.5, 1.1, 0.8, and 0.5.}
\scalebox{0.85}{
\resizebox{0.82\textwidth}{!}{
\begin{tabular}{c ccccc ccc}
\hline
&\multicolumn{8}{c}{\textbf{M1}} \\ 
\hline
   &5\% & 25\% & 50\% & 75\% & 95\% &$\mathcal{P}_{d_1} (\%)$& $\mathcal{P}_{d_2}(\%)$ &$\mathcal{P}_{d_3}(\%)$ \\
\hline
\textbf{SIS}&10.0&10.0&10.0&10.0&10.0       
&100.0&100.0&100.0    \\
\textbf{BanditSIS ($\alpha=1.5$)}&10.0&10.0&10.0&10.0&894.0  &89.80&94.00&95.20   \\
\textbf{BanditSIS ($\alpha=1.1$)}&10.0&10.0&10.0&10.0&188.3  &97.20&97.80&98.60   \\
\textbf{BanditSIS ($\alpha=0.8$)}&10.0&10.0&10.0&10.0&10.0   &99.80&99.80&99.80   \\
\textbf{BanditSIS ($\alpha=0.5$)}&10.0&10.0&10.0&10.0&10.0   &100.0&100.0&100.0   \\
\hline

&\multicolumn{8}{c}{\textbf{M2}} \\ 
\hline
  &5\% & 25\% & 50\% & 75\% & 95\% &$\mathcal{P}_{d_1}(\%)$& $\mathcal{P}_{d_2}(\%)$ & $\mathcal{P}_{d_3}(\%)$ \\
\hline
\textbf{SIS}&10.0&10.0&10.0&11.0&471.3       
&92.20&93.40&94.60    \\
\textbf{BanditSIS ($\alpha=1.5$)}&10.0&10.0&61.0&1336.0&2000.0   &60.00&65.60&69.60  \\
\textbf{BanditSIS ($\alpha=1.1$)}&10.0&10.0&10.0&599.0&2000.0   &71.40&75.00&76.80   \\
\textbf{BanditSIS ($\alpha=0.8$)}&10.0&10.0&10.0&87.0&2000.0   
&78.20&80.00&82.80   \\
\textbf{BanditSIS ($\alpha=0.5$)}&10.0&10.0&10.0&11.0&2000.0   
&85.00&86.80&87.80   \\
\hline

&\multicolumn{8}{c}{\textbf{M3}} \\ 
\hline
  &5\% & 25\% & 50\% & 75\% & 95\% &$\mathcal{P}_{d_1}(\%)$& $\mathcal{P}_{d_2}(\%)$ & $\mathcal{P}_{d_3}(\%)$ \\
\hline
\textbf{SIS}&10.0&10.0&10.0&10.0&10.0       
&99.80&100.0&100.0    \\
\textbf{BanditSIS ($\alpha=1.5$)}&10.0&10.0&271.0&894.0&2000.0   &57.60&68.20&74.40   \\
\textbf{BanditSIS ($\alpha=1.1$)}&10.0&10.0&10.0&402.0&2000.0   &77.00&82.80&85.40   \\
\textbf{BanditSIS ($\alpha=0.8$)}&10.0&10.0&10.0&10.0&1336.0   &90.60&92.40&93.40   \\
\textbf{BanditSIS ($\alpha=0.5$)}&10.0&10.0&10.0&10.0&10.0   &98.40&98.40&98.80   \\
\hline

&\multicolumn{8}{c}{\textbf{M4}} \\ 
\hline
  &5\% & 25\% & 50\% & 75\% & 95\% &$\mathcal{P}_{d_1}(\%)$& $\mathcal{P}_{d_2}(\%)$ & $\mathcal{P}_{d_3}(\%)$ \\
\hline
\textbf{SIS}&10.0&10.0&10.0&10.0&10.0       
&100.0&100.0&100.0    \\
\textbf{BanditSIS ($\alpha=1.5$)}&10.0&10.0&184.0&451.3&1336.0   &66.60&80.40&86.20   \\
\textbf{BanditSIS ($\alpha=1.1$)}&10.0&10.0&10.0&126.0&916.1   &84.60&90.80&93.60   \\
\textbf{BanditSIS ($\alpha=0.8$)}&10.0&10.0&10.0&10.0&402.0   &96.80&97.80&98.80   \\
\textbf{BanditSIS ($\alpha=0.5$)}&10.0&10.0&10.0&10.0&10.0   &99.60&100.0&100.0   \\
\hline
\end{tabular}
}}
\end{table}

The $\mathcal{S}$ is used to measure the precision of ranking the predictors of a specified screening procedure. It corresponds to the lowest rank among all active predictors after the procedure has ranked the features in terms of importance. The closer $\mathcal{S}$ is to the number of active features, (i.e., $dt=10$ in $\textbf{M1}$--$\textbf{M4}$), the more effective this screening procedure is. 
The $\mathcal{P}$ is an indicator for accuracy such that as the estimated model size $d$ becomes sufficiently large, the probability $\mathcal{P}$ approaches one. Here we set $d_1 = \lfloor n/\log(n) \rfloor$, $d_2 = 2\lfloor n/\log(n) \rfloor$, $d_3 = 3\lfloor n/\log(n) \rfloor$ to empirically assess the impact of the cutoff during our simulations.

Table \ref{tab:res} summarizes the results of SIS and BanditSIS under the two evaluation criteria across settings $\textbf{M1}$--$\textbf{M4}$. 
We observe that as the hyperparameter $\alpha$ decreases, the performance of BanditSIS progressively aligns with that of SIS in terms of both evaluation metrics $\mathcal{S}$ and $\mathcal{P}$.
In particular, when $\alpha \leq 0.8$, the differences between the results of two methods become negligible in most settings, indicating the convergence of BanditSIS's accuracy toward that of SIS.
Taking into account both the accuracy of the results and the computational efficiency, a conservative and practical choice for $\alpha$ lies in the range $0.5 \leq \alpha \leq 1.1$.
This range strikes a favorable balance; smaller values of $\alpha$ correspond to larger subsample sizes in each iteration, resulting in higher accuracy of results.
These empirical findings provide intuitive support for Theorem \ref{thm: sure_screening}, which quantitatively characterizes how the hyperparameter $\alpha$ influences the algorithmic accuracy.
In addition, the results indicate that BanditSIS adaptively adjusts the structure of the screening procedure and substantially reduces unnecessary computations related to non-informative features.

\begin{figure}[h!t] 
    \centering 
    \includegraphics[width=0.4\textwidth]{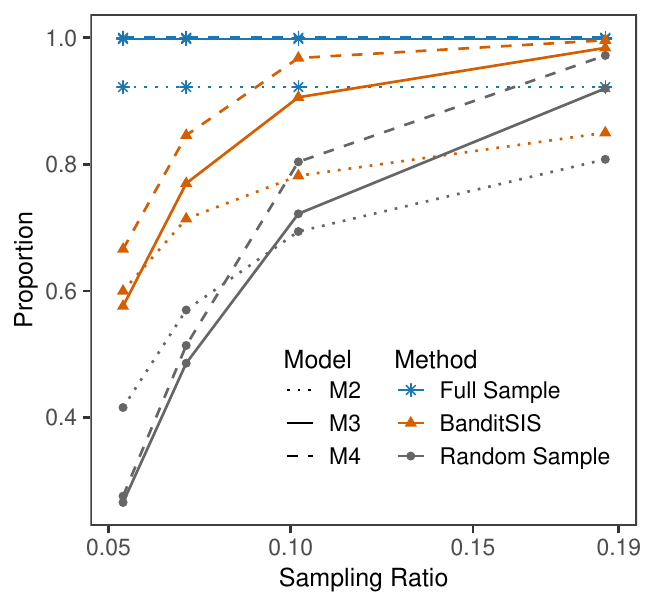} 
    \caption{The comparison of BanditSIS, SIS with random sampling w.r.t. the sampling ratio versus the proportion $\mathcal{P}$ with $d = d_1$ for $\textbf{M2}-\textbf{M4}$ over 500 replications. The sampling ratios (from left to right) correspond to BanditSIS with values of hyperparameter $\alpha = $1.5, 1.1, 0.8 and 0.5. For reference, SIS with the full sample is included as horizontal blue lines.} 
    \label{fig: ratio_proportion}
\end{figure}

Figure~\ref{fig: ratio_proportion} illustrates the relationship between the proportion $\mathcal{P}$ over 500 repetitions and varying sampling ratios, to demonstrate the performance improvement achieved by BanditSIS compared to SIS under an equal-probability random sampling strategy. For reference, the performance of SIS using the full sample is also included in the figure.
From left to right, the sampling ratios correspond to those used by BanditSIS with different values of the hyperparameter $\alpha = 1.5, 1.1, 0.8$ and 0.5, respectively.
We observe that BanditSIS consistently outperforms SIS with random sampling, particularly when the sampling ratio is low (i.e., larger $\alpha$).
This performance gain highlights the strength of our adaptive sampling strategy, which is especially beneficial in scenarios involving large datasets where efficient dimensionality reduction is crucial.
In contrast, as the sampling ratio increases (that is, $\alpha$ decreases), the performance gap between BanditSIS and random sampling gradually narrows.
This outcome is expected, as the advantage of adaptive subsampling diminishes when a larger fraction of the data is utilized.
Based on these observations, we recommend choosing the hyperparameter $\alpha$ in the range $0.5 \leq \alpha \leq 1.1$ for practical applications, with further adjustment depending on computational budget and desired accuracy.

\subsection{Computational time}
\begin{figure}[h!t]
    \captionsetup{width=1\linewidth}
    \centering
    \subfigure[Sample size $n$ vs. CPU time.]{
    \includegraphics[height=5.5cm]{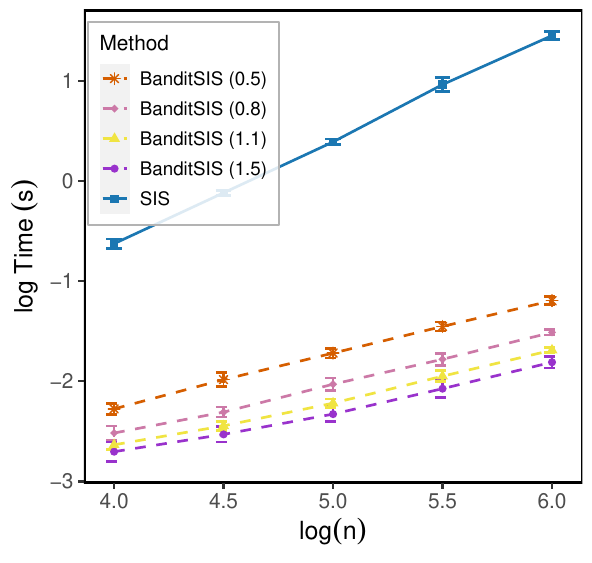}\label{fig: time_compared_n}
    }
    \subfigure[Feature number $p$ vs. CPU time.]{
    \includegraphics[height=5.5cm]{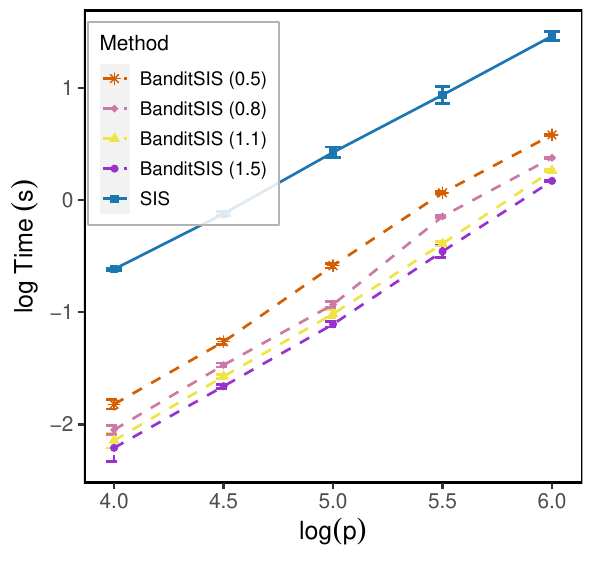}\label{fig: time_compared_p}
    }
    \caption{(a) The sample size $n$ versus the CPU time (in seconds) of SIS and BanditSIS for different values of hyperparameter $(\alpha = 1.5, 1.1, 0.8\text{ and }0.5)$ with $p=2000, d=30$. (b) The number of features $p$ versus the CPU time (in seconds) of SIS and BanditSIS for the values of hyperparameter $(\alpha = 1.5, 1.1, 0.8\text{ and }0.5)$ with $n=2000, d=30$. Vertical bars are the standard errors w.r.t. 100 replications.}\label{fig: time_compared}
\end{figure}

We compare the average CPU time (in seconds) vs. the sample size $n$ and the feature number $p$ for SIS and BanditSIS with different values of the hyperparameter $\alpha$ to screen a specified number $d$ features over 100 replications. Here we set $d=30$, the number of features $p=2000$ in the first case, and the sample size $n=2000$ in the second case. 
The results are shown in Fig.~\ref{fig: time_compared}. The magnitude of the slope of the curve in the graph of $\log(\text{Time})$ versus $\log(n)$ or $\log(p)$ measures how quickly the computational time of the algorithms grows with the sample size $n$ (on the left) or the feature number $p$ (on the right).
As shown in Fig.~\ref{fig: time_compared_n}, SIS requires more CPU time, and its time significantly increases with sample size compared to BanditSIS. 
This observation is expected since the computational cost of SIS is $O(np)$, while that of BanditSIS is $O(\sqrt{n}p)$. Such a result indicates that BanditSIS is more efficient for screening non-informative features, especially when the sample size $n$ is extremely large.
We also observe that BanditSIS requires more CPU time when $\alpha$ decreases. This observation is consistent with the second term of computational complexity, $O(\sqrt{n}p/\alpha^2)$, which indicates that smaller values of $\alpha$ lead to the selection of more subsamples per iteration, which in turn yields more accurate estimates of the correlation coefficients.
In other words, the hyperparameter $\alpha$ governs a trade-off between the computational efficiency and the performance of the proposed BanditSIS algorithm.
Fig.~\ref{fig: time_compared_p} illustrates that the computation time of both SIS and BanditSIS grows rapidly with the number of features $p$. This is because while redundancy is reduced in feature screening, BanditSIS retains a linear dependence on $p$.
We also observe that the behavior of BanditSIS is closer to SIS with a smaller value of $\alpha$, which can be explained by the nature of the subsample function in Equation~(\ref{eqn:nl}).

\subsection{Analysis of a real dataset}
We consider the MovieLens 25M Dataset,
% \footnote{The MovieLens 25M dataset used in this study was created by GroupLens Research, and is openly available at \url{https://files.grouplens.org/datasets/movielens/ml-25m-README.html}.}, 
which was generated in November 2019 by the MovieLens movie recommendation service \citep{harper2015movielens}.
This dataset contains 25 million ratings and tag genome data with 15 million relevance scores. 
These data were amassed from the contributions of over 100,000 users between January 1995 and November 2019. They are widely used in machine learning, data mining, and recommendation systems research \citep{wu2021self,lin2022improving,wu2022graph,hou2024large}.

After preprocessing data, we retained 13,815 movies, each associated with relevance scores under more than 630,000 tags. 
Following the procedure in \citet{amagata2021reverse} and \citet{wu2024accelerating}, we selected 100 representative users from a pool of 162,541 users and used the the matrix factorization technique in \citet{chin2016libmf} to estimate each user's ratings for the aforementioned movies. 
We define the response variable as a representative user's ratings for all 13,815 movies, and the features are the relevance scores across 636,756 tags. Thus, the dataset has the sample size $n = 13,815$ and $p = 636,756$ features.
For each representative user, we randomly partition 70\% of the data as the training set, treating the rest as the testing set, and keeping $d = 100$ features in the screening procedures. We use the features we retained to fit a multiple linear regression.
Finally, we compare the mean absolute prediction errors and computational times of SIS and BanditSIS with different values of \(\alpha\) over the 100 representative users.
The results presented in Table~\ref{table: real_data} show that SIS requires considerable computational time to screen out features compared to BanditSIS. Notably, the proposed BanditSIS method achieves a prediction error comparable to SIS while requiring nearly a hundredth of its CPU time.

\begin{table*}[h!t]
\centering
\caption{\label{table: real_data} Mean absolute errors of prediction error and time consumption (in seconds) based on SIS and BanditSIS with different values of hyperparameter ($\alpha=1.5$, $1.1$, and $0.5$) for 70\% training set over the 100 representative users.}
\scalebox{0.85}{
\begin{tabular}{c cccc}
\hline
\multirow{2}{*}{} & \multirow{2}{*}{\textbf{SIS}}
&\textbf{BanditSIS}&\textbf{BanditSIS}&\textbf{BanditSIS} \\
&&$\alpha=1.5$&$\alpha=1.1$ &$\alpha=0.5$\\
\hline
$R^2$ & 0.5727 ($\pm$0.1782) &0.5706 ($\pm$0.1806)&0.5702 ($\pm$0.1813)&0.5717 ($\pm$0.1807)\\
Mean Absolute Error &0.3400 ($\pm$0.1330) &0.3407 ($\pm$0.1335)&0.3406 ($\pm$0.1338)&0.3402 ($\pm$0.1337)\\
Time (s) &11718.9&227.2&272.9&637.2\\
\hline
\end{tabular}
}
\end{table*}

%%%%%%%%%%%%%%%%%\bibliography{ref}%%%%%%%%%%%%%%%%%%%%%%%%%%%%%%%%%%%%%%%%%%%%%%%%%%%%%%%%%%%%%%%%%%%%%%%%%a
%                                                                                        %
%                                                                                        %
%                                                                                        %
%                                                                                        %
%                                                                                        %
%                                                                                        %
%                                     Section 6                                          %
%                                                                                        %
%                                                                                        %
%                                                                                        %
%                                                                                        %
%                                                                                        %
%                                                                                        %
%%%%%%%%%%%%%%%%%%%%%%%%%%%%%%%%%%%%%%%%%%%%%%%%%%%%%%%%%%%%%%%%%%%%%%%%%%%%%%%%%%%%%%%%%%
\section{Conclusion}\label{sec:conclusion}
In this work, we proposed BanditSIS, an adaptive subsampling method for sure independence screening, especially when the sample size and feature dimension are extremely large. Motivated by the successive elimination idea from MAB learning, by regarding each feature as an arm and the marginal Pearson correlation as the reward, we significantly reduce the computational complexity from $O(np)$ to $O(\sqrt{n}p)$ while preserving the well-known sure screening property under mild regularity conditions. 
Numerical studies show that BanditSIS achieves screening and prediction performance comparable to SIS while requiring much less computational time.
Extending our framework to non-linear and model-free feature screening algorithms \citep{li2012feature,zhu2011model,li2023scalable}, as well as integrating it with modern subsampling strategies \citep{meng2017effective,li2020modern,li2024core}, will be a focus of our future work.

\section*{Acknowledgment}
This work is supported by National Key R\&D Program of China No. 2023YFC3304701.

\bibliographystyle{chicago}
\bibliography{ref}

@article{donoho2000high,
  title={High-dimensional data analysis: {T}he curses and blessings of dimensionality},
  author={Donoho, David L and others},
  journal={AMS Math Challenges Lecture},
  volume={1},
  number={2000},
  pages={32},
  year={2000}
}

@inproceedings{fan2006statistical,
  title={Statistical challenges with high dimensionality: {F}eature selection in knowledge discovery},
  author={Fan, Jianqing and Li, Runze},
  booktitle={25th International Congress of Mathematicians, ICM 2006},
  year={2006}
}

@incollection{hotelling1992relations,
  title={Relations between two sets of variates},
  author={Hotelling, Harold},
  booktitle={Breakthroughs in Statistics: Methodology and Distribution},
  pages={162--190},
  year={1992},
  publisher={Springer}
}

@article{fisher1936use,
  title={The use of multiple measurements in taxonomic problems},
  author={Fisher, Ronald A},
  journal={Annals of Eugenics},
  volume={7},
  number={2},
  pages={179--188},
  year={1936},
  publisher={Wiley Online Library}
}

@article{mead1992review,
  title={Review of the development of multidimensional scaling methods},
  author={Mead, Al},
  journal={Journal of the Royal Statistical Society: Series D (The Statistician)},
  volume={41},
  number={1},
  pages={27--39},
  year={1992},
  publisher={Wiley Online Library}
}

@article{borg2003modern,
  title={Modern Multidimensional Scaling: {T}heory and Applications},
  author={Borg, I and Groenen, P},
  journal={Journal of Educational Measurement},
  volume={40},
  number={3},
  pages={277--280},
  year={2003},
  publisher={Wiley}
}

@article{roweis2000nonlinear,
  title={Nonlinear dimensionality reduction by locally linear embedding},
  author={Roweis, Sam T and Saul, Lawrence K},
  journal={Science},
  volume={290},
  number={5500},
  pages={2323--2326},
  year={2000},
  publisher={American Association for the Advancement of Science}
}

@article{zhang2004principal,
  title={Principal manifolds and nonlinear dimensionality reduction via tangent space alignment},
  author={Zhang, Zhenyue and Zha, Hongyuan},
  journal={SIAM Journal on Scientific Computing},
  volume={26},
  number={1},
  pages={313--338},
  year={2004},
  publisher={SIAM}
}

@article{hinton2002stochastic,
  title={Stochastic neighbor embedding},
  author={Hinton, Geoffrey E and Roweis, Sam},
  journal={Advances in Neural Information Processing Systems},
  volume={15},
  pages={833--840},
  year={2002}
}

@article{tibshirani1996regression,
  title={Regression shrinkage and selection via the lasso},
  author={Tibshirani, Robert},
  journal={Journal of the Royal Statistical Society Series B: Statistical Methodology},
  volume={58},
  number={1},
  pages={267--288},
  year={1996},
  publisher={Oxford University Press}
}

@article{fan2001variable,
  title={Variable selection via nonconcave penalized likelihood and its oracle properties},
  author={Fan, Jianqing and Li, Runze},
  journal={Journal of the American Statistical Association},
  volume={96},
  number={456},
  pages={1348--1360},
  year={2001},
  publisher={Taylor \& Francis}
}

@article{efron2004least,
  title={Least Angle Regression},
  author={Efron, Bradley and Hastie, Trevor and Johnstone, Iain and Tibshirani, Robert},
  year={2004},
  journal={Annals of Statistics},
  pages={407--451},
  publisher={JSTOR}
}

@article{candes2007dantzig,
  title={The {D}antzig selector: {S}tatistical estimation when p is much larger than n},
  author={Candes, Emmanuel and Tao, Terence},
  year={2007},
  journal={Annals of Statistics},
  volume={35},
  number={1},
  pages={2313--2351}
}

@article{zou2006adaptive,
  title={The adaptive lasso and its oracle properties},
  author={Zou, Hui},
  journal={Journal of the American Statistical Association},
  volume={101},
  number={476},
  pages={1418--1429},
  year={2006},
  publisher={Taylor \& Francis}
}

@article{zou2005regularization,
  title={Regularization and variable selection via the elastic net},
  author={Zou, Hui and Hastie, Trevor},
  journal={Journal of the Royal Statistical Society Series B: Statistical Methodology},
  volume={67},
  number={2},
  pages={301--320},
  year={2005},
  publisher={Oxford University Press}
}

@article{zou2009adaptive,
  title={On the adaptive elastic-net with a diverging number of parameters},
  author={Zou, Hui and Zhang, Hao Helen},
  journal={Annals of Statistics},
  volume={37},
  number={4},
  pages={1733--1751},
  year={2009},
  publisher={NIH Public Access}
}

@article{liu2018feature,
  title={Feature selection of gene expression data for cancer classification using double {R}{B}{F}-kernels},
  author={Liu, Shenghui and Xu, Chunrui and Zhang, Yusen and Liu, Jiaguo and Yu, Bin and Liu, Xiaoping and Dehmer, Matthias},
  journal={BMC Bioinformatics},
  volume={19},
  number={1},
  pages={1--14},
  year={2018},
  publisher={BioMed Central}
}

@article{fan2009ultrahigh,
  title={Ultrahigh dimensional feature selection: {B}eyond the linear model},
  author={Fan, Jianqing and Samworth, Richard and Wu, Yichao},
  journal={The Journal of Machine Learning Research},
  volume={10},
  pages={2013--2038},
  year={2009},
  publisher={JMLR. org}
}

@article{fan2008sure,
  title={Sure independence screening for ultrahigh dimensional feature space},
  author={Fan, Jianqing and Lv, Jinchi},
  journal={Journal of the Royal Statistical Society Series B: Statistical Methodology},
  volume={70},
  number={5},
  pages={849--911},
  year={2008},
  publisher={Oxford University Press}
}

@article{chakrabarti2008mortal,
  title={Mortal multi-armed bandits},
  author={Chakrabarti, Deepayan and Kumar, Ravi and Radlinski, Filip and Upfal, Eli},
  journal={Advances in Neural Information Processing Systems},
  volume={21},
  year={2008}
}

@article{berry1985bandit,
  title={Bandit problems: {S}equential allocation of experiments (Monographs on statistics and applied probability)},
  author={Berry, Donald A and Fristedt, Bert},
  journal={London: Chapman and Hall},
  volume={5},
  number={71-87},
  pages={7--7},
  year={1985},
  publisher={Springer}
}

@article{even2006action,
  title={Action Elimination and Stopping Conditions for the Multi-Armed Bandit and Reinforcement Learning Problems.},
  author={Even-Dar, Eyal and Mannor, Shie and Mansour, Yishay and Mahadevan, Sridhar},
  journal={The Journal of Machine Learning Research},
  volume={7},
  number={6},
  pages={1079--1105},
  year={2006}
}

@article{kaufmann2016complexity,
  title={On the complexity of best arm identification in multi-armed bandit models},
  author={Kaufmann, Emilie and Capp{\'e}, Olivier and Garivier, Aur{\'e}lien},
  journal={The Journal of Machine Learning Research},
  volume={17},
  pages={1--42},
  year={2016}
}

@inproceedings{ding2013multi,
  title={Multi-armed bandit with budget constraint and variable costs},
  author={Ding, Wenkui and Qin, Tao and Zhang, Xu-Dong and Liu, Tie-Yan},
  booktitle={Proceedings of the AAAI Conference on Artificial Intelligence},
  volume={27},
  pages={232--238},
  year={2013}
}

@inproceedings{tran2010epsilon,
  title={Epsilon-first policies for budget-limited multi-armed bandits},
  author={Tran-Thanh, Long and Chapman, Archie and De Cote, Enrique Munoz and Rogers, Alex and Jennings, Nicholas R},
  booktitle={Proceedings of the AAAI Conference on Artificial Intelligence},
  volume={24},
  pages={1211--1216},
  year={2010}
}

@inproceedings{jamieson2014best,
  title={Best-arm identification algorithms for multi-armed bandits in the fixed confidence setting},
  author={Jamieson, Kevin and Nowak, Robert},
  booktitle={2014 48th Annual Conference on Information Sciences and Systems (CISS)},
  pages={1--6},
  year={2014},
  organization={IEEE}
}

@article{gabillon2012best,
  title={Best arm identification: {A} unified approach to fixed budget and fixed confidence},
  author={Gabillon, Victor and Ghavamzadeh, Mohammad and Lazaric, Alessandro},
  journal={Advances in Neural Information Processing Systems},
  volume={25},
  year={2012}
}

@article{chen2014combinatorial,
  title={Combinatorial pure exploration of multi-armed bandits},
  author={Chen, Shouyuan and Lin, Tian and King, Irwin and Lyu, Michael R and Chen, Wei},
  journal={Advances in  Neural Information Processing Systems},
  volume={27},
  year={2014}
}

@article{liu2022model,
  title={Model-free feature screening and {F}{D}{R} control with knockoff features},
  author={Liu, Wanjun and Ke, Yuan and Liu, Jingyuan and Li, Runze},
  journal={Journal of the American Statistical Association},
  volume={117},
  number={537},
  pages={428--443},
  year={2022},
  publisher={Taylor \& Francis}
}

@article{li2012feature,
  title={Feature screening via distance correlation learning},
  author={Li, Runze and Zhong, Wei and Zhu, Liping},
  journal={Journal of the American Statistical Association},
  volume={107},
  number={499},
  pages={1129--1139},
  year={2012},
  publisher={Taylor \& Francis}
}

@inproceedings{liu2019bandit,
  title={A bandit approach to maximum inner product search},
  author={Liu, Rui and Wu, Tianyi and Mozafari, Barzan},
  booktitle={Proceedings of the AAAI Conference on Artificial Intelligence},
  volume={33},
  pages={4376--4383},
  year={2019}
}

@article{harper2015movielens,
  title={The movielens datasets: {H}istory and context},
  author={Harper, F Maxwell and Konstan, Joseph A},
  journal={ACM Transactions on Interactive Intelligent Systems},
  volume={5},
  number={4},
  pages={1--19},
  year={2015},
  publisher={Acm New York, NY, USA}
}

@inproceedings{lin2022improving,
  title={Improving graph collaborative filtering with neighborhood-enriched contrastive learning},
  author={Lin, Zihan and Tian, Changxin and Hou, Yupeng and Zhao, Wayne Xin},
  booktitle={Proceedings of the ACM Web Conference 2022},
  pages={2320--2329},
  year={2022}
}

@article{wu2022graph,
  title={Graph neural networks in recommender systems: {A} survey},
  author={Wu, Shiwen and Sun, Fei and Zhang, Wentao and Xie, Xu and Cui, Bin},
  journal={ACM Computing Surveys},
  volume={55},
  number={5},
  pages={1--37},
  year={2022},
  publisher={ACM New York, NY}
}

@inproceedings{hou2024large,
  title={Large language models are zero-shot rankers for recommender systems},
  author={Hou, Yupeng and Zhang, Junjie and Lin, Zihan and Lu, Hongyu and Xie, Ruobing and McAuley, Julian and Zhao, Wayne Xin},
  booktitle={European Conference on Information Retrieval},
  pages={364--381},
  year={2024},
  organization={Springer}
}

@article{wu2021self,
  title={Self-supervised learning on graphs: {C}ontrastive, generative, or predictive},
  author={Wu, Lirong and Lin, Haitao and Tan, Cheng and Gao, Zhangyang and Li, Stan Z},
  journal={IEEE Transactions on Knowledge and Data Engineering},
  volume={35},
  number={4},
  pages={4216--4235},
  year={2021},
  publisher={IEEE}
}

@article{chin2016libmf,
  title={L{I}{B}{M}{F}: {A} library for parallel matrix factorization in shared-memory systems},
  author={Chin, Wei-Sheng and Yuan, Bo-Wen and Yang, Meng-Yuan and Zhuang, Yong and Juan, Yu-Chin and Lin, Chih-Jen},
  journal={Journal of Machine Learning Research},
  volume={17},
  number={86},
  pages={1--5},
  year={2016}
}

@inproceedings{amagata2021reverse,
  title={Reverse maximum inner product search: {H}ow to efficiently find users who would like to buy my item?},
  author={Amagata, Daichi and Hara, Takahiro},
  booktitle={Proceedings of the 15th ACM Conference on Recommender Systems},
  pages={273--281},
  year={2021}
}

@article{wu2024accelerating,
  title={Accelerating Matrix Factorization by Dynamic Pruning for Fast Recommendation},
  author={Wu, Yining and Duan, Shengyu and Sai, Gaole and Cao, Chenhong and Zou, Guobing},
  journal={arXiv preprint arXiv:2404.04265},
  year={2024},
  pages={arXiv--2404}
}

@article{li2023scalable,
  title={Scalable model-free feature screening via sliced-{W}asserstein dependency},
  author={Li, Tao and Yu, Jun and Meng, Cheng},
  journal={Journal of Computational and Graphical Statistics},
  volume={32},
  number={4},
  pages={1501--1511},
  year={2023},
  publisher={Taylor \& Francis}
}

@article{li2024core,
  title={Core-elements for large-scale least squares estimation},
  author={Li, Mengyu and Yu, Jun and Li, Tao and Meng, Cheng},
  journal={Statistics and Computing},
  volume={34},
  number={6},
  pages={190},
  year={2024},
  publisher={Springer}
}

@incollection{meng2017effective,
  title={Effective statistical methods for big data analytics},
  author={Meng, Cheng and Wang, Ye and Zhang, Xinlian and Mandal, Abhyuday and Zhong, Wenxuan and Ma, Ping},
  booktitle={Handbook of research on applied cybernetics and systems science},
  pages={280--299},
  year={2017},
  publisher={IGI Global}
}

@article{li2020modern,
  title={Modern subsampling methods for large-scale least squares regression},
  author={Li, Tao and Meng, Cheng},
  journal={International Journal of Cyber-Physical Systems (IJCPS)},
  volume={2},
  number={2},
  pages={1--28},
  year={2020},
  publisher={IGI Global}
}

@article{zhu2011model,
  title={Model-free feature screening for ultrahigh-dimensional data},
  author={Zhu, Li-Ping and Li, Lexin and Li, Runze and Zhu, Li-Xing},
  journal={Journal of the American Statistical Association},
  volume={106},
  number={496},
  pages={1464--1475},
  year={2011},
  publisher={Taylor \& Francis}
}

@article{li2012robust,
  title={Robust rank correlation based screening},
  author={Li, Gaorong and Peng, Heng and Zhang, Jun and Zhu, Lixing},
  journal={The Annals of Statistics},
  volume={40},
  number={3},
  pages={1846--1877},
  year={2012},
  publisher={Citeseer}
}

@article{cui2015model,
  title={Model-free feature screening for ultrahigh dimensional discriminant analysis},
  author={Cui, Hengjian and Li, Runze and Zhong, Wei},
  journal={Journal of the American Statistical Association},
  volume={110},
  number={510},
  pages={630--641},
  year={2015},
  publisher={Taylor \& Francis}
}
\end{document}